# Investigating Spatiotemporal Redundancy in Video Transformer for Collision Anticipation

Xiaoshan Zhou

School of Project Management, Faculty of Engineering, University of Sydney, NSW, Australia, xiaoshan.zhou@sydney.edu.au

**Abstract**

In worker–equipment proximity monitoring and safe human–robot interaction, video transformers are increasingly used for collision anticipation and have demonstrated strong predictive performance. However, their accuracy comes with substantial computational demands, creating a tension with the need for low-latency inference on mobile robotic platforms and the pursuit of lower-carbon computation in construction. To address this problem, rather than designing a new lightweight architecture, this study investigates where computation within an established video transformer is redundant and whether that redundancy can be removed without materially degrading predictive performance. Using VideoMAEv2-Base on the Nexar Collision Prediction dataset, we first examine how collision-relevant information evolves across network depth and then investigate two complementary forms of redundancy: structured capacity redundancy in multilayer perceptrons (MLPs) and spatiotemporal redundancy in the token stream. Linear probes show that interpretable motion cues, including flow magnitude, looming, and approach versus retreat, are most accessible at intermediate layers, whereas collision-label discrimination strengthens toward the final layer. Token redundancy is strongly axis-specific: adjacent temporal-token similarity reaches 0.970 in later layers, while spatial similarity falls to 0.297, indicating substantially greater redundancy across time than across space. Exploiting this asymmetry, temporal token merging reduces backbone computation from 356.99 to 178.50 GFLOPs and latency from 12.69 to 7.21 ms per clip, a 1.76× speedup, while mean average precision changes only from 0.7478 to 0.7443. Importance-guided retention of 50% of MLP units preserves an AUC of 0.753, compared with 0.529 under matched random retention, and reveals that pruning alters score calibration before discriminative ranking collapses. These findings provide the first empirical evidence of network- and token-level spatiotemporal redundancy in a video transformer for collision anticipation and establish a new evidence-based pathway for pursuing faster and more computationally sustainable algorithms through targeted temporal token compression and neuron pruning.

Keywords: construction safety; video transformer; collision anticipation; structured pruning; token compression

## 1 Introduction

Despite long-standing safety interventions, construction continues to experience more than 1,000 occupational fatalities annually in the United States (CPWR 2025). In 2024, there were approximately 163,600 nonfatal injuries, corresponding to 2.1 injuries per 100 full-time workers (BLS 2025). Struck-by incidents remain the leading cause of nonfatal construction injuries and the second-leading cause of fatalities (NIOSH 2023). Among struck-by hazards, half are related to motorized vehicles or powered machinery (NIOSH 2023), such as excavators, loaders, trucks, and forklifts. The safety concerns become even more serious looking forward, as more jobsite robotics are introduced (79% of contractors reported using jobsite robotics in 2026, compared with 29% in 2025; BuiltWorlds 2026), NIOSH explicitly identifies collisions between robots and workers as an emerging construction safety concern (NIOSH 2024), and, compared with manufacturing, where robots run in controlled designated areas, the dynamic and interactive nature of a collaborative co-robot setup (Liu and Jebelli 2024; Gao and Adel 2026) makes collision anticipation more pressing.

Struck-by hazards have long been studied in worker–equipment proximity monitoring (Ding et al. 2026; Kudlinski and Ryu 2026). Early radio-frequency systems detected when a worker entered an equipment safety zone and issued a warning (Teizer et al. 2010). Subsequent systems used 3D position, heading, and speed to reduce the false alarms of simple proximity thresholds (Wang and Razavi 2016), while Kim et al. (2016) introduced vision-based methods to assess the spatial relation between worker and equipment from site imagery, and Tang et al. (2020) tested trajectory forecasting from real construction site videos. At the same time, methodologically, deep learning methods have evolved from long short-term memory (LSTM) models (Cai et al. 2020) to transformer-based predictions (Alduais et al. 2026; Wang et al. 2026), which continue to be the dominant method (Chu et al. 2026).

Despite the high performance, video transformers introduce very high computation, as each frame contributes spatial patches, longer or higher-resolution clips expand the spatiotemporal token stream, attention operates across tokens, and token-wise projections and multilayer perceptrons (MLPs) add work at every layer (Wang et al. 2023). There is a growing trend revealing that not every visual token, frame, or layer is necessary, and exploring more efficient machine vision perception for low latency and low carbon emission (Bourzac 2024; Wang et al. 2025). Researchers have explored lightweight network architectures for edge computing and deployment on robotic platforms in construction (Chang et al. 2023; Xiao et al. 2025), but it is reported that in construction safety risk assessment, achieving higher accuracy needs to sacrifice computing efficiency (He et al. 2025).

This study aims to seek an alternative way to address this problem, based on the empirical findings from previous works that a substantial amount of neurons, connections, and attention heads in neural networks can be removed with limited loss on vision tasks. For example, AlexNet could be compressed by about 35× and VGG-16 by about 49× without loss of ImageNet accuracy (Han et al. 2015). Frankle and Carbin (2019) proposed “the lottery ticket hypothesis” and found that 80% sparsity in ResNet-50 could still reach the full network’s ImageNet accuracy. Voita et al. (2019) studied individual attention heads in transformer machine translation and found that removing 79% of the heads results in only a 0.15 BLEU drop. In the video domain, transformer cost becomes particularly severe because the temporal dimension dramatically increases the token sequence (Selva et al. 2023), and evidence about temporal and inter-frame information redundancy has been found to guide pruning with very minor accuracy drops (Ding et al. 2023; Chen et al. 2025). But current studies primarily establish redundancy for video action recognition, with no network- or token-level redundancy evidence for collision anticipation with a lead-time window. In addition, current studies do not separate temporal similarity from spatial similarity, so we do not know whether spatiotemporal redundancy holds asymmetrically, and whether adjacent frames contain almost the same scene so that they can be compressed, or whether there is redundant spatial information such that we may need focused attention while overlooking the irrelevant locations in the scene. A systematic MLP pruning and video-token merging study is able to give us answers to these questions. This is the first study of its kind in video for structured model compression to lower inference cost without erasing the evolving motion cues that distinguish dangerous convergence from harmless proximity.

Specifically, we first characterize how collision-relevant information is reorganized in transformer layers across depth. Observable motion variables computed independently from the network, including approach versus retreat, radial looming, ego-expansion, flow magnitude, and temporal-difference energy, are used as probes of what can be linearly recovered from frozen intermediate representations. We then distinguish two forms of redundancy that have different computational consequences: structured capacity redundancy in the transformer MLPs and spatiotemporal redundancy in the token stream. The former is tested by importance-guided versus random unit retention; the latter is tested by random removal, redundancy-based dropping, attention-based merging, and matched temporal versus spatial merging under controlled token schedules and FLOP budgets. Taken together, we aim to answer systematically along which axis, and how much, computation can be reduced in a video transformer for collision anticipation without damaging its accuracy, so as to preserve algorithm progress in the research community while seeking lighter computation.

## 2 Literature Review

### 2.1 Worker-Vehicle Collision Anticipation

Earlier proximity systems in construction make the struck-by risk observable with cameras, radio tags, ultra-wideband, or GPS, and warn when worker–equipment separation falls below a threshold (Teizer et al. 2010; Zhu et al. 2016; Wang and Razavi 2016; Bang et al. 2021). Yet identical distances can describe either an approaching or a retreating vehicle, which pose significantly different risk levels, and localization error and occlusion often challenge the reliability of a geometric boundary. Collision anticipation therefore moves forward to take into account not only proximity, but also the direction and development of motion, that is, trajectory forecasting. For example, Tang et al. (2020) extract worker and equipment trajectories from site video and use a multihead model to forecast locations at several future times. Cai et al. (2020) add workers' individual movement and contextual cues from nearby agents, work groups, and likely destinations to an LSTM trajectory model. Zhang and Ge (2022) replaced the LSTM with a transformer to track crane loads and workers in camera video and predict their paths for dynamic prewarning. Jeon et al. (2024) introduce prediction-assisted MPSORT to recover noisy multicamera tracks of workers and vehicles and project forecasts onto a bird's-eye site map. Wang et al. (2026) integrate YOLOv11 and SORT tracking with transformer trajectory forecasts and dynamic risk regions reflecting motion and positional uncertainty. Transformer-based approaches have been found dominant for trajectory forecasting in construction (Alduais et al. 2026), and has been empirically reported to substantially outperform recurrent sequence models (Lu and Ergan 2026). However, despite the high accuracy, its computational request is rarely evaluated in the literature.

To reduce computational burden, earlier work such as Chen et al. (2022) studied the feasibility of edge computing for vision-based construction resource detection. Xiao et al. (2025) sought to design a lightweight detector for edge deployment on a mobile robot device. The limitation of these studies is that researchers need to craft new network architectures for lighter design (Zhang et al. 2024; Zhang 2026), then the large body of work reflecting algorithmic progress for prediction performance becomes separate and non-useable. In our study, instead, we investigate a better way to reduce computation that still leverages the published literature on algorithm design, but upon that, seeks to find redundancy within the architecture, then we can use MLP pruning or token merging to reduce the model to a much smaller size while preserving its established accuracy.

### 2.2 Network Pruning and Video-Token Reduction

Large neural networks can contain substantial task-dependent redundancy, but the way redundancy is removed determines whether computation actually falls. Weight pruning creates sparse connections, structured pruning removes entire units, heads, or layers, and token reduction shortens the sequence processed by later blocks, lowering both token-wise projections and the quadratic attention term (Han et al. 2015; Molchanov et al. 2019; Rao et al. 2021). These mechanisms have different effects on memory, latency, and the evidence available to a collision head.

Pruning studies show that some trained capacity can be removed, but also delimit what can be inferred from that result. Han et al. (2015) reduced AlexNet from 61 million to 6.7 million weights and VGG-16 from 138 million to 10.3 million without an accuracy loss after pruning and retraining. Frankle and Carbin (2019) showed that small, trainable subnetworks exist within larger initialized networks, and thus proposed the lottery-ticket hypothesis. SparseGPT achieves at least 50% weight sparsity in a large language model (LLM) without retraining (Frantar and Alistarh 2023), whereas ShortGPT removes 10 of 40 layers from LLaMA-2-13B and changes MMLU from 55.0 to 52.2 (Men et al. 2025). Sandri et al. (2025) remove feed-forward neurons and examine 25%, 37.5%, and 50% structural sparsity. These precedents motivate our MLP-unit ranking, but their language and static-image outcomes do not establish pre-impact video-warning performance.

For token reduction, image transformers provide an analogous choice between retaining salient tokens and aggregating apparently redundant ones. DynamicViT learns to discard low-scoring patches progressively (Rao et al. 2021); EViT retains highly attended patches and fuses less attended patches (Liang et al. 2022). Token Merging uses pairwise similarity and a merge budget without retraining the backbone (Bolya et al. 2023). Dropping loses the removed token, while averaging can carry coarse context forward but may dilute a small moving object. These understanding is conceptual, but we do not yet know whether attention or cosine similarity can identify and preserve all task-critical collision evidence.

Video adds a distinct temporal axis to this problem, i.e., reducing tokens from within a frame to across frames, or along both axes (Hwang et al. 2024). The training-free vid-TLDR method uses attention-derived saliency to distinguish important patches and applies saliency-aware reduction and merging to video transformers (Choi et al. 2024). Pollard and Wray (2025) vary training-free similarity-based video merging across TimeSformer, Motionformer, VideoMAE, and ViViT on action recognition datasets. DyCoke merges redundant information across frames and selectively reduces spatial key–value cache tokens during video and LLM inference for video question answering (QA) and video captioning (Tao et al. 2025). FrameFusion compares spatially corresponding tokens from adjacent frames, merges similar ones in early layers, and then prunes less important tokens in deeper layers for video retrieval and QA (Fu et al. 2025). The anchor and optimal transport method of J. Li et al. (2026) transfers information from tokens that would otherwise be removed into retained tokens within and across frames for video QA. These studies establish the existence of spatiotemporal redundancy in video transformers, but their tasks are action recognition, retrieval, and video QA. There is no evidence signaling how redundancy flows in video transformer for collision anticipation.

**3 Methodology3.1 VideoMAEv2 Backbone and Study Design**

VideoMAEv2 has become a common backbone for models designed for video understanding and prediction, such as VAGNet (Vipulananthan and Chitraranjan 2026), AdaTAD (Liu et al. 2024), and SurgFUTR (Sharma et al. 2025), and it has reported the highest AP on several collision anticipation datasets (S. Li et al., 2026).

VideoMAEv2 is a self-supervised video representation backbone, and its design is the outcome of several successive developments. It starts from the Vision Transformer (ViT), which, instead of using convolutions, divides an image into patches, converts each patch into a token, adds positional information, and processes the token sequence with transformer blocks (Dosovitskiy et al. 2021). The next development was the masked autoencoder (MAE), which masks about 75% of the image patches during pretraining and uses a much smaller decoder to reconstruct the missing pixels (He et al. 2022). VideoMAE extended the MAE from a 2D image patch to a 3D video cube, or tubelet, and found that the model can still learn effectively when 90–95% of the video tokens are masked, because neighboring frames are highly correlated (Tong et al. 2022). Related variants refine which tokens are masked or retained, for example through motion-guided masking (Huang et al. 2023) and the removal of redundant spatiotemporal tokens during pretraining (Hwang et al. 2024). Although the encoder had become much lighter, the decoder remained computationally heavy because it must reconstruct the whole video token space. VideoMAEv2 introduces dual masking, in which the encoder still uses high-ratio tube masking while the decoder uses running-cell masking to reconstruct a selected subset of the masked video cubes instead of processing every possible token (Wang et al. 2023).

For this study, we used VideoMAEv2-Base with 86 million parameters. A 16 × 224 × 224 clip is embedded with 2 × 16 × 16 tubelets into 8 temporal bins and a 14 × 14 spatial grid, giving 1,568 tokens of width 768. VideoMAEv2-Base has 12 transformer blocks, each with 12 attention heads and a feed-forward MLP expansion ratio of four. The final tokens are mean-pooled and passed through LayerNorm(768), dropout(0.1), and a two-class linear head. Token reduction changes the sequence entering later attention and MLP computations, whereas MLP-unit suppression changes a fixed portion of the feed-forward computation without shortening the token sequence (Wang et al. 2023; Bolya et al. 2023).

We fine-tuned the model on our dataset for three epochs with batch size 1 and gradient accumulation over 24 steps, AdamW (learning rate $10^{-5}$; weight decay 0.05), cosine annealing, cross-entropy loss, bfloat16 autocasting, and gradient clipping at 5. Horizontal flipping with probability 0.5 was used for augmentation. The checkpoint maximizing validation mean AP across the three anticipation horizons was retained. All subsequent analyses use this frozen backbone; prediction heads are refitted for the compressed-token experiments.

Two computational resources, the token stream and the feed-forward capacity, organize our experiments, and the complete study design is summarized in Figure 1. Because the backbone is frozen after checkpoint selection, every difference reported later is attributable to the intervention rather than to retraining. Section 3.3 extracts pooled representations at selected depths and defines the observable motion variables that serve as probe targets. Section 3.4 specifies the linear probes that measure how much of each variable, and of the collision label, remains linearly recoverable at each depth. Section 3.5 characterizes the geometry of the same representations through effective dimensionality and through separate temporal and spatial token similarity, which identifies the axis along which redundancy accumulates. Section 3.6 then intervenes on the token stream, comparing random removal, redundancy-based dropping, attention-based merging, and matched temporal versus spatial merging under controlled token schedules and FLOP budgets. Section 3.7 intervenes on the feed-forward capacity instead, suppressing MLP hidden units ranked by first-order importance and comparing this with matched random suppression. Throughout, matched random, schedule, and compression-axis controls separate the selection rule from the amount and location of computation removed (Molchanov et al. 2019; Wang et al. 2023; Bolya et al. 2023). All computation ran on an NVIDIA H200 GPU with Python 3.12.9.

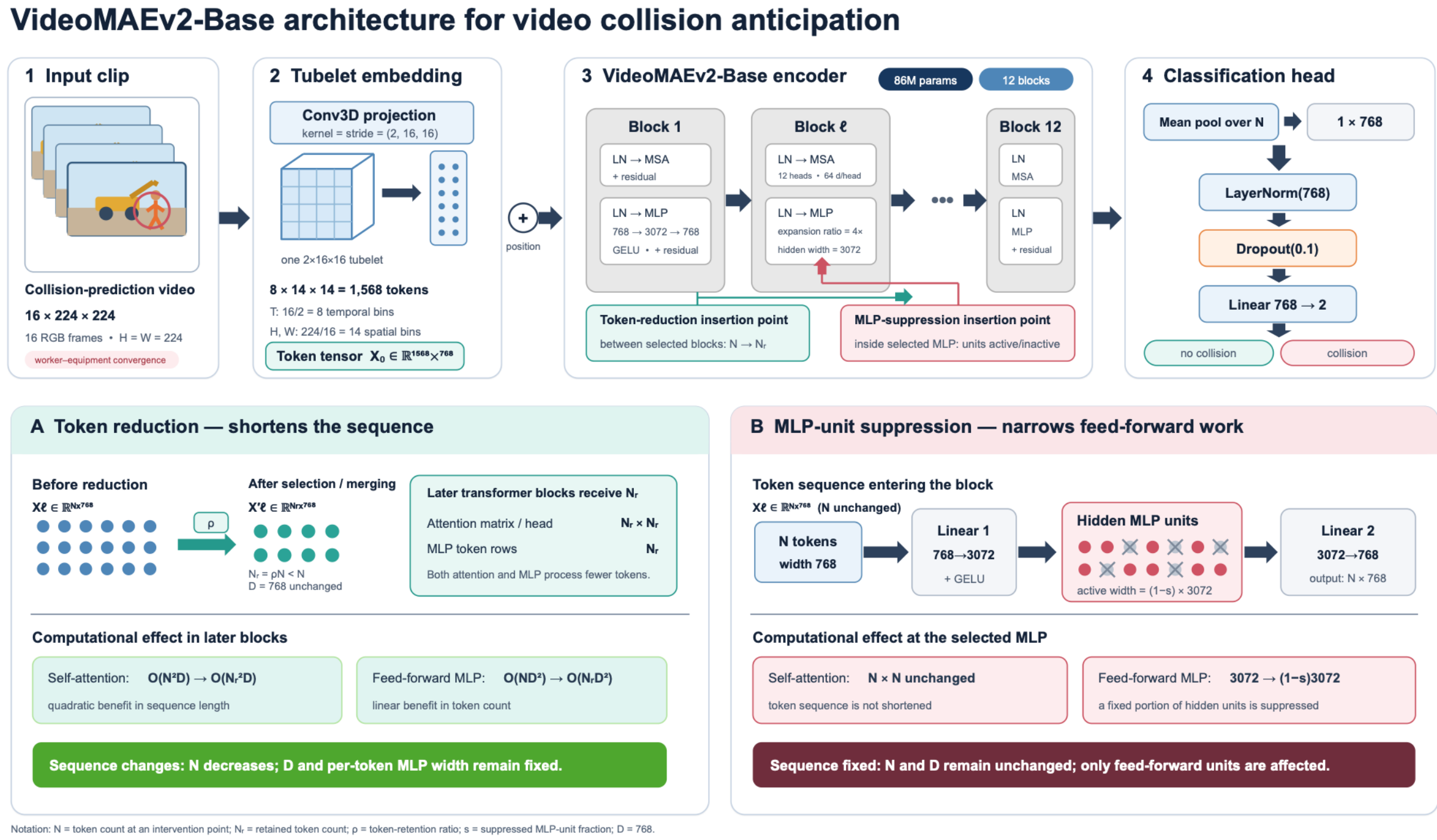

Figure 1. Overview of VideoMAEv2 and compression methods. Depth probes and token geometry diagnose the intact backbone; token reduction is applied after selected blocks, while MLP-unit suppression targets feed-forward activations in all blocks.

### 3.2 Dataset and Preprocessing

The Nexar Collision Prediction dataset (Moura et al. 2025) was used in our experiment, because it uniquely provides collision anticipation rather than merely accident recognition. Table 1 compares it with another nine public dashcam video datasets that are widely used for accident anticipation, detection, and causality analysis. Most of these resources were designed to recognize or localize an accident once it has become visible. For example, DAD places the accident in the final 10 frames of each clip (Chan et al. 2016); A3D and DoTA annotate the start and the end of an anomaly (Yao et al. 2019, 2023); CTA labels a cause interval and an effect interval (You and Han 2020); DADA-2000 defines an appearance-based accident window (Fang et al. 2022); ROL marks the appearance of the risky object (Karim et al. 2023); and CTAD supplies simulation-generated temporal ground truth (Luo and Wang 2023). CCD and AoTA do provide exact accident-frame labels (Bao et al. 2020; Guan et al. 2025), but neither marks the moment at which a developing hazard first becomes recognizable.

Table 1. Public dashcam video datasets for collision and accident anticipation, detection, and analysis.

| Dataset | Main purpose | Videos / clips | Frames / duration | Positive vs. negative | Main temporal structure | Environment / visual diversity |
|---|---|---|---|---|---|---|
| Nexar (Moura et al. 2025) | Collision anticipation | 1,500 | 1.70M; ~40 s; ~30 fps | 750 positive + 750 negative; positives = 400 collisions + 350 near-collisions | Alert time + exact event time | Weather, lighting, road type; fog, wet roads, blur, lens flare, reflections, extreme viewpoints |
| DAD (Chan et al. 2016) | Accident anticipation | 1,750 clips | 100 frames; 5 s; 20 fps | 620 accident + 1,130 normal | Accident is placed in final 10 frames | Real crowded Taiwan streets; multiple road users, but environmental metadata relatively limited |
| A3D (Yao et al. 2019) | Accident/anomaly detection | 1,500 accident clips | 128,175 frames; 23–208 frames; 10 fps | Primarily accident clips; normal training data drawn separately from HEV-I | Inevitable-anomaly start + anomaly end | Sunny/rainy/snowy; urban/countryside; diverse viewpoints |
| CTA (You and Han 2020) | Accident causality recognition | 1,935 | 9.53 h total | Accident videos | Cause interval + effect/collision interval | Web-sourced; mostly dashcams, ~12% monitoring cameras |
| CCD (Bao et al. 2020) | Accident anticipation | 4,500 | 50 frames; 5 s; 10 fps | 1,500 accidents + 3,000 normal | Exact accident-frame labels | Day/night; normal/rain/snow; ego involvement labeled |
| DADA-2000 (Fang et al. 2022) | Accident understanding + driver attention | 2,000 | 658,476; 30 fps | Accident videos | Appearance-based accident window | Highway/urban/rural/tunnel; sunny/rainy/snowy; day/night |
| DoTA (Yao et al. 2023) | Traffic anomaly detection | 4,677 original release | 731,932; 10 fps | Anomaly videos; pre-anomaly portions supply normal observations | Inevitable-anomaly start + anomaly end | Different weather and lighting; real-world dashcams |
| ROL (Karim et al. 2023) | Risky-object localization / early accident analysis | 1,000 | 100 frames; 5 s; 20 fps | Accident videos | Risky-object appearance + exact accident beginning | Road type/intersection, weather, lighting |
| CTAD | Urban accident | 1,100 | ~727K | Accident + normal | Simulation- | Controllable simulated |

| Dataset | Main purpose | Videos / clips | Frames / duration | Positive vs. negative | Main temporal structure | Environment / visual diversity |
|---|---|---|---|---|---|---|
| (Luo and Wang 2023) | detection | | frames | traffic scenes | generated temporal GT | conditions |
| AoTA (Guan et al. 2025) | Accident anticipation | 4,800 | 50 frames; 10 fps | 3,600 accident + 1,200 normal | Precisely reannotated accident occurrence | Wider driving environments; weather and day/night labels |

We used 316 videos as the training partition for backbone fine-tuning and 110 videos as the validation partition for checkpoint selection, thresholding, and calibration. We kept a further 110 videos as a seen test partition to support early model comparisons, and the remaining 964 videos for the exploratory representation analysis and the compression experiments.

Representation probes use five-fold cross-validation grouped by parent video. Before the later compression results were computed, video identifiers were hashed and stratified by label and scene into a 485-video development split and a 479-video reserve split. The main pruning and token results use the 485-video development split.

Clips were decoded with the same ffmpeg settings for every condition. Each clip has 16 frames at 8 fps, spanning 1.875 s, resized to 224 × 224 pixels without aspect-ratio preservation. Pixel values were normalized as $(x/255 - 0.5)/0.5$. For positives, a clip ends $h \in \{0.5, 1.0, 1.5\}$ s before the annotated event, excluding the collision frame. Negative videos use a pseudo-event time to prevent window position alone from revealing the class. Training windows were sampled with negative undersampling, whereas evaluation uses complete, unmasked clips at all three horizons (0.5 s, 1 s, and 1.5 s).

### 3.3 Layerwise Representation Analysis

Before removing any units or tokens, we characterize what different transformer depths encode and how redundancy develops. Specifically, we test whether observable collision-relevant motion cues, including approach versus retreat, residual radial looming, flow magnitude, ego-expansion, and temporal change, are accessible from the learned video representation, and whether their accessibility changes as features are progressively transformed toward the final collision decision. This analysis also establishes the representational structure of the uncompressed model before the subsequent pruning and token-compression experiments.

For transformer block $l$, let

$$H^{(l)} \in \mathbb{R}^{N \times d} \quad (1)$$

denote the token representation produced by that block, where $N = 1{,}568$ is the number of spatiotemporal tokens and $d = 768$ is the embedding dimension. To obtain a clip-level representation, the token embeddings were mean-pooled:

$$z^{(l)} = \frac{1}{N} \sum_{i=1}^{N} H_i^{(l)}, \quad (2)$$

where $z^{(l)} \in \mathbb{R}^{768}$ summarizes the clip at depth $l$. We extracted representations from Blocks 0, 2, 4, 6, 8, 10, and 11, thereby sampling the progression from early to intermediate and final transformer representations without requiring analysis of every block.

The probe targets introduced above are five observable properties of the input clip: temporal change, flow magnitude, ego-expansion, looming, and approach versus retreat. They are ordered from generic frame-to-frame change to the relational quantity that defines a collision threat, and all of them are computed directly from the decoded pixels, so that probe performance reflects what is recoverable from the learned representation rather than a quantity that the network itself produced (Kowal et al. 2024). Let $I_t$ denote the grayscale frame at time t for t = 1, …, T, let Ω be the set of pixel positions, and let $f_t(p)$ be the dense optical flow at position p between frames t and t + 1.

Temporal change is the simplest cue and is quantified as the temporal-difference energy

$$E = \frac{1}{(T-1)|\Omega|} \sum_{t,p} \left(I_{t+1}(p) - I_t(p)\right)^2 \quad (3)$$

which records how much the image changes from frame to frame without distinguishing what moved or in which direction. It is included because it bounds from below the dynamic information that any layer should retain, and because a representation that loses it has lost the rawest form of motion evidence.

Flow magnitude summarizes the amount of apparent motion irrespective of its direction,

$$M = \frac{1}{(T-1)|\Omega|} \sum_{t,p} \|f_t(p)\|_2 \tag{4}$$

Unlike temporal-difference energy, it separates displacement from photometric change and therefore remains interpretable under illumination variation, but it still carries no information about whether the motion is approaching or receding.

The remaining three variables exploit the radial structure of the flow field, because on a forward-facing camera a closing object appears as expansion about the point toward which the camera is moving. With c denoting the image center and $r(p) = (p - c)/\|p - c\|_2$ the unit radial direction at position p, the radial flow component is

$$\lambda_t(p) = \langle f_t(p), r(p) \rangle \tag{5}$$

where $\langle \cdot, \cdot \rangle$ is the inner product. The quantity $\lambda_t(p)$ is positive when a point moves away from the image center, that is, when its projection expands.

Ego-expansion is the part of that radial field attributable to the motion of the camera itself. It was estimated over the peripheral region $\Omega_{per}$, where the scene is dominated by static background swept outward by forward ego-motion:

$$G = \frac{1}{(T-1)|\Omega_{\mathrm{per}}|} \sum_{t,p \in \Omega_{\mathrm{per}}} \lambda_t(p) \tag{6}$$

Ego-expansion is retained as a separate target because a clip can show strong global expansion simply because the vehicle is travelling quickly, with no object on a collision course; separating it prevents speed from being mistaken for threat.

Looming is the residual radial expansion in the central region $\Omega_{cen}$ once the ego-expansion component has been removed, and is referred to throughout as the residual radial looming:

$$L = \frac{1}{(T-1)|\Omega_{\mathrm{cen}}|} \sum_{t,p \in \Omega_{\mathrm{cen}}} (\lambda_t(p) - G) \tag{7}$$

Subtracting $G$ is what makes $L$ a hazard-related rather than a speed-related quantity: it isolates the part of central expansion that cannot be explained by the observer's own forward motion, which is the optical signature of an object closing on the camera.

Approach versus retreat is the binary form of the same residual, with $y_{\text{app}} = 1$ when $L > 0$ and 0 otherwise. It is included as a categorical target because the sign of the residual, rather than its magnitude, separates a dangerous convergence from a harmless recession, and because a representation can encode the amount of motion accurately while losing its direction. All five variables were computed once per clip at the decoding resolution used for the backbone, so each clip carries one value per target.

### 3.4 Linear Probes

We used linear probes to determine which observable properties of the video could be recovered from each frozen transformer depth. A linear probe fits a simple prediction model to $\boldsymbol{z}^{(l)}$ while keeping the VideoMAEv2 backbone unchanged. High probe performance indicates that information about a variable is linearly accessible from that layer. The probe targets are the five motion variables defined in Eqs. (3)–(7) together with the collision label. The continuous targets were the ego-expansion rate, the residual radial looming, the flow magnitude, and the temporal-difference energy, which capture different aspects of scene motion, from overall camera expansion to residual radial motion and generic frame-to-frame change. The categorical targets were the collision label and approach versus retreat, the latter being the sign of the residual radial looming and therefore representing whether the relevant motion pattern was converging or receding.

To prevent information from the same source video appearing in both training and evaluation, all probes used five-fold cross-validation grouped by video. All clips corresponding to the different anticipation horizons of a given video were assigned to the same fold. This ensures that probe performance reflects generalization across videos rather than recognition of video-specific appearance.

For a continuous target, the feature matrix from a given layer was standardized using statistics estimated from the training portion of each fold, and the target was mean-centered. Ridge regression was then fitted by minimizing

$$\hat{\beta} = \underset{\beta}{\operatorname{argmin}}[\| y - X\beta \|_2^2 + \alpha \| \beta \|_2^2], \tag{8}$$

where $X$ contains the standardized pooled layer representations, $y$ is the corresponding mean-centered continuous target, and $\alpha$ controls the amount of $L_2$ regularization. We considered

$$\alpha \in \{1,10,100,1000\}, \tag{9}$$

and selected the value according to out-of-fold predictive performance within the probe procedure. Performance for continuous variables was reported using the coefficient of determination, $R^2$.

For categorical targets, ridge-regularized logistic regression was fitted to the same frozen layer representations. The classifier estimates

$$P\big(y = 1 \mid z^{(l)}\big) = \sigma\big(\beta_0 + \beta^\top z^{(l)}\big), \tag{10}$$

where $\sigma(\cdot)$ is the logistic function. We used $\alpha = 100$, with optimization performed for 60 Newton iterations, and evaluated out-of-fold discrimination using the area under the receiver operating characteristic curve (AUC).

### 3.5 Representation Dimensionality and Token Redundancy

This subsection examines how redundant the transformer's representations become across network depth. This analysis addresses a different question from the linear probes, as the latter asks whether a variable such as approach/retreat or collision status can be recovered from a representation, while the present analysis asks how much independent representational variation remains, and whether redundancy emerges preferentially across time or across space.

We considered two measures of redundancy. The first is the effective rank, which summarizes how many independent dimensions meaningfully contribute to a representation (Roy and Vetterli 2007). Let the singular values of a representation matrix be $\sigma_i$. We first convert the squared singular values into normalized variance weights,

$$p_i = \frac{\sigma_i^2}{\sum_j \sigma_j^2}, \tag{11}$$

and compute the entropy-based effective rank as

$$r_{\text{eff}} = \exp(-\textstyle\sum_i p_i \log p_i). \tag{12}$$

A high effective rank indicates that variance is distributed across many dimensions, whereas a low effective rank indicates that the representation is concentrated in a smaller number of dominant dimensions and is therefore more redundant. We additionally report the participation ratio,

$$r_{\text{PR}} = \frac{\left(\sum_i \sigma_i^2\right)^2}{\sum_i \sigma_i^4}, \tag{13}$$

and $n_{90}$, defined as the minimum number of principal components required to explain 90% of the variance.

Two forms of effective rank were analyzed because they answer different questions. Pooled-feature effective rank was computed across clips using the matrix of clip-level pooled representations $z^{(l)}$. It measures how many dimensions are used to distinguish different videos at a particular transformer depth. Token effective rank was computed within each clip using the centered $1{,}568 \times 768$ token matrix. This quantity measures how diverse the individual spatiotemporal tokens remain inside a single video. A layer can become more discriminative across videos while simultaneously containing increasingly redundant tokens within each video.

Effective rank describes overall dimensionality but does not indicate where the redundancy occurs. We therefore separately measured similarity along the temporal and spatial axes. For each clip, the 1,568 tokens were reshaped according to their original tubelet arrangement into

$$8 \text{ temporal bins} \times 196 \text{ spatial positions} \times 768 \text{ dimensions}. \quad (14)$$

To quantify temporal redundancy, the 196 spatial token embeddings within each temporal bin b were first averaged to obtain a temporal summary vector $g_b \in \mathbb{R}^{768}$. After $L_2$ normalization, cosine similarity was computed between consecutive temporal bins and averaged across the seven adjacent pairs:

$$C_{\text{temp}} = \frac{1}{7}\sum_{b=1}^{7}\left\langle \frac{g_b}{\|g_b\|_2}, \frac{g_{b+1}}{\|g_{b+1}\|_2} \right\rangle. \quad (15)$$

where $\langle \cdot,\cdot \rangle$ denotes the inner product. Values approaching one indicate that neighboring temporal representations have become highly similar, suggesting that later processing may be repeatedly operating on nearly duplicated temporal information.

To quantify spatial redundancy, the 196 spatial tokens within each temporal bin were individually $L_2$-normalized. We then calculated the mean pairwise cosine similarity between all distinct spatial-token pairs and averaged the result across the eight temporal bins:

$$C_{\text{spat}} = \frac{1}{8}\sum_{b=1}^{8}\frac{\sum_{m\neq n}\langle \hat{h}_{b,m}, \hat{h}_{b,n}\rangle}{196\times 195}, \quad (16)$$

where $h_{b,m}$ denotes the $L_2$-normalized embedding of the spatial token at position m in temporal bin b. A high $C_{\text{spat}}$ indicates that different spatial locations have become similar in representation space, whereas lower values indicate that spatial positions remain differentiated.

Both temporal and spatial similarity measures were computed on the same fixed, stratified subsample of 240 clips. We separate these two axes because temporally adjacent observations may become highly similar while spatial locations remain distinguishable, or vice versa. This hypothesis is also tested later using matched temporal and spatial compression interventions rather than inferred from representation statistics alone.

### 3.6 Token-Compression Methods

After characterizing how token similarity changes across transformer depth, we next tested whether this redundancy could be exploited to reduce computation. The central question is whether the way tokens are selected or merged matters for preserving collision-relevant information. The methods compared below instantiate the strategies that the video-token-reduction literature has established as the main alternatives, namely attention- or saliency-based retention, similarity-based merging within and across frames, and redundancy-adaptive layerwise schedules (Ding et al. 2023; Feng et al. 2024; Chen et al. 2025). We additionally include random token removal as the baseline. The comparisons are matched for computational budget and, where required, for the number of tokens retained at each layer. All token-reduction methods operate inside the frozen VideoMAEv2 backbone without retraining its transformer weights. Reduction is introduced at predefined transformer depths, and a new prediction head is fitted on the resulting frozen representations.

To compare methods at matched computational cost, we analytically estimated the floating-point operations (FLOPs) of each transformer block from its number of input tokens $n$ and embedding width $d = 768$. Counting one multiply–add as two operations, the approximate block cost is

$$F(n) = 24nd^2 + 4n^2d. \tag{17}$$

The first term represents the token-wise linear operations: query, key, and value projections, the output projection, and the two MLP transformations with an expansion ratio of four. The second term represents the two quadratic attention operations, namely computation of the attention logits and multiplication of the attention weights by the value vectors. For the dense model, each block processes $n = 1{,}568$ tokens, giving a total cost across the 12 transformer blocks of approximately

$$12F(1568) = 356.99 \text{ GFLOPs}. \tag{18}$$

Token reduction was applied after Blocks 2, 4, 6, 8, and 10. This follows previous studies in which more tokens are retained in early layers and pruning becomes more aggressive once later representations have become more focused (Chen et al. 2025). We evaluated four target computational ratios relative to dense inference: $\rho \in \{0.80, 0.70, 0.60, 0.50\}$. More aggressive FLOP targets were excluded. Because the first three transformer blocks remain dense, they alone impose a lower bound of approximately $3/12 = 25\%$ of dense transformer computation. In preliminary schedule calculations, target ratios of 0.35 or lower reduced the later representations to only 8–22 surviving tokens. Such settings would test near-collapse of the token representation rather than meaningful redundancy removal. The final operating grid was therefore fixed at $\rho = 0.50$–$0.80$.

For the five reduction stages, the fraction of tokens retained at stage $j$ was parameterized as

$$\text{keep}_j = r^{w_j}, \tag{19}$$

where $w_j$ specifies the desired pruning profile and the scalar $r$ was solved numerically by bisection so that the resulting whole-backbone computation matched the target FLOP ratio $\rho$ to within 0.1%. This formulation guarantees that every stage retains a valid fraction in (0, 1].

Two schedules were evaluated. The fixed schedule used $w = (1,1,1,1,1)$, so that compression pressure was distributed uniformly across the reduction stages. The depth-aware schedule used $w = (0.40, 0.70, 1.00, 1.30, 1.60)$, thus applying relatively mild compression at earlier layers and progressively stronger compression at later layers.

When tokens are merged, the resulting token inherits the temporal and spatial indices of its dominant parent, while its size is updated to represent the combined number of original tokens. Every token carries a size $s_i$, and the final clip representation is computed using size-weighted pooling. If the retained token set is $\mathcal{K}$,

$$z = \frac{\sum_{i \in \mathcal{K}} s_i x_i}{\sum_{i \in \mathcal{K}} s_i}. \tag{20}$$

where $x_i$ is the embedding of token i. This ensures that a merged token contributes in proportion to the amount of original information it represents. In the absence of merging, all $s_i = 1$, and the expression reduces exactly to the ordinary mean pooling used for the dense model. This common pooling rule keeps dropping- and merging-based methods on a comparable representational scale.

*Random token removal*

The baseline, denoted R, removes tokens without considering their content. At each reduction stage, every token receives an independent random score, the required $k$ tokens are retained, and the remainder are discarded. Because random selection can itself produce substantial performance variability, each random configuration was evaluated with three independent pruning seeds.

*Redundancy-based token dropping*

The S method tests whether tokens can be removed according to representational redundancy. For each token $i$, we identify its eight most similar neighboring tokens in representation space and form their size-weighted mean. Redundancy is then defined as the cosine similarity between token $i$ and this local representative:

$$\mathrm{red}(i) = \cos\left(x_i, \ \mathrm{norm}\left[\frac{\sum_{j \in NN_8(i)} s_j x_j}{\sum_{j \in NN_8(i)} s_j}\right]\right). \tag{21}$$

where $NN_8(i)$ is the set of the eight nearest neighbors of token i in representation space. A large value indicates that the token is highly similar to information already represented by neighboring tokens, so tokens with the highest redundancy scores are removed first. The neighborhood is defined in representation space rather than on the original image grid to make the criterion applicable even after earlier compression stages have disrupted the regular spatiotemporal arrangement of tokens.

*Attention-based selection with token merging*

The V method represents an attention-based compression strategy. It assigns importance according to how much attention each token receives across attention heads and query positions. The score for token *i* is

$$\text{score}(i) = s_i \frac{1}{H} \sum_{h=1}^{H} \sum_q \text{softmax}\left(\frac{Q_h K_h^\top}{\sqrt{d_h}}\right)_{q,i}, \tag{22}$$

where $H$ is the number of attention heads, $d_h$ is the dimensionality of each head, and $Q_h$ and $K_h$ are the query and key matrices of head h.

The $k$ highest-scoring tokens are retained. In addition, instead of discarding the remaining tokens, each discarded token is assigned to its most similar retained token and incorporated through a size-weighted merge. Thus, V combines attention-based token importance with representation-preserving aggregation.

*Temporal and spatial neighbor merging*

The T and Tsp methods test whether the axis along which tokens are compressed matters. For temporal merging (T), tokens are first grouped by spatial position and then ordered by temporal index. Tokens that correspond to the same spatial location at neighboring temporal positions can therefore be paired. For spatial merging (Tsp), tokens are instead grouped by temporal position and ordered by spatial index, allowing neighboring spatial tokens within the same temporal slice to be paired.

Within each ordered group, alternating positions form destination and source sets. Candidate source–destination pairs are valid only when they share the non-merged coordinate: temporal pairs must correspond to the same spatial position, while spatial pairs must belong to the same temporal bin. Among the valid pairs, cosine similarity is used to identify the most redundant neighbors.

To reach a target of $k$ tokens from $N$ current tokens, the $N - k$ most similar valid pairs are merged first. For destination token $x_{\text{dst}}$ and source token $x_{\text{src}}$, with sizes $s_{\text{dst}}$ and $s_{\text{src}}$, respectively, the merged representation is

$$x_{\text{new}} = \frac{s_{\text{dst}} x_{\text{dst}} + s_{\text{src}} x_{\text{src}}}{s_{\text{dst}} + s_{\text{src}}}, \tag{23}$$

with updated size

$$s_{\text{new}} = s_{\text{dst}} + s_{\text{src}}. \tag{24}$$

As a result, the most similar neighboring tokens along the chosen axis are combined first, while their aggregate contribution to later pooling is preserved.

Two controls separate the token criterion from its schedule and axis. $R_{depth}$ uses the same depth-aware token counts as the informed arms but chooses tokens at random. $\text{Tsp}_{depth}$ matches $T_{depth}$ in schedule, per-layer token counts, and estimated FLOPs, changing the merge axis from temporal to spatial. These controls are necessary because a performance difference at equal final token count could otherwise arise from different early-layer compute or selection pressure (Bolya et al. 2023; Pollard and Wray 2025).

The first control, $R_{depth}$, uses exactly the same depth-aware token schedule as the guided depth-aware methods but selects retained tokens randomly. $R_{depth}$ therefore matches the number of tokens processed at every layer and the resulting FLOPs, while removing any informed selection criterion. Comparing a guided depth-aware method with $R_{depth}$ isolates the contribution of which tokens are selected from the contribution of the depth schedule itself.

The second control, $\text{Tsp}_{depth}$, is designed specifically to test the temporal-versus-spatial redundancy hypothesis. It uses the same depth-aware schedule, retained token counts, and FLOP budget as temporal merging, but performs merging along the spatial rather than the temporal axis. It therefore holds the computational budget and layer schedule constant while changing only the compression axis.

*Prediction heads and random seeds*

Because a powerful classifier head could partially compensate for damage caused by token compression, every compressed representation was evaluated using two prediction heads. The linear head provides a relatively strict diagnostic of representation quality because it has limited capacity to reorganize a degraded feature space. The MLP-768 head provides a higher-capacity readout and tests the predictive performance achievable from the compressed representation.

Each head was fitted using three independent training seeds. Weight decay was selected from validation performance averaged across the three head seeds, and all three fitted models were retained for evaluation rather than selecting the best seed. Random token-pruning conditions additionally used three independent pruning seeds to quantify the variability introduced by stochastic token selection.

### 3.7 Structured MLP Pruning

We separately test feed-forward capacity with first-order Taylor unit importance, which scores an activation by the sensitivity of the positive-class logit to suppressing that activation (Molchanov et al. 2019). Each of the 12 blocks contains 3,072 MLP hidden units. Guided and three-seed random controls suppress the same number of post-GELU unit activations within each block.

Each of the 12 transformer blocks contains an MLP with 3,072 hidden units. To rank these units by task relevance, we used a first-order Taylor importance criterion, following the general principle of estimating how much the model output would change if a component were removed (Molchanov et al. 2019). For hidden unit u, let $a_u$ denote its post-GELU activation and $z_+$ the positive-class collision logit. Importance was accumulated over C = 48 calibration clips as

$$\mathrm{Imp}(u) = \sum_{c=1}^{C} \left| \mathrm{mean}_{b,i} \left( a_u \cdot \frac{\partial z_+}{\partial a_u} \right) \right| \tag{25}$$

where the inner mean is taken over the batch and token dimensions of each calibration clip. This score combines the magnitude of a unit's activation with the sensitivity of the collision logit to that activation, so a unit receives a high score when it is both active and influential on the model's positive collision prediction.

Within each transformer block, units were ranked by this score and the top $k$ were retained, where

$$k = \mathrm{round}(r \times 3072), \tag{26}$$

and $r$ denotes the MLP-unit retention ratio. The remaining units were suppressed by setting their post-GELU activations to zero through forward hooks. This procedure preserves the original trained weights and isolates the effect of removing unit activations without retraining the backbone.

To determine whether performance depends on which units are retained rather than simply on reducing MLP capacity, we included a matched random-pruning control. For each retention ratio, the same number of units was selected uniformly at random within each block, with three independent random seeds. We evaluated retention ratios of

$$r \in \{1.0,\ 0.8,\ 0.6,\ 0.5,\ 0.4,\ 0.3,\ 0.2\}. \tag{27}$$

MLP-unit retention is distinct from whole-model parameter retention: at 50% unit retention, the corresponding share of model parameters associated with retained structures is approximately 67.15%. Our forward-hook intervention zeros activations but leaves the original dense matrices allocated and executed. Parameter fractions therefore indicate potential structural removal; they are not measured MLP-pruning latency, memory, or energy savings.

Ranking and threshold behavior were evaluated separately because pruning may compress the logit scale before changing the ordering of examples. We report AP and AUC, balanced accuracy at both a fixed 0.5 threshold and one selected on validation data, logit standard deviation, and expected calibration error (Guo et al. 2017). The validation-selected threshold was applied unchanged to evaluation clips.

As a validation-only calibration analysis, Platt scaling was fitted as

$$p_{\text{cal}} = \sigma(az_{+} + b) \tag{28}$$

where a and b are scalar parameters fitted on the validation predictions only and then held fixed for evaluation, and $z_+$ is the uncalibrated positive-class logit defined above. This post hoc calibration tests whether shifted scores can be corrected without using evaluation labels (Guo et al. 2017).

## 4 Results

The results follow the analysis sequence. We first examine how dynamic collision cues and representation geometry change across VideoMAEv2 depth, then test axis-specific token compression and its efficiency–performance trade-offs, and finally evaluate structured MLP pruning and calibration effects.

### 4.1 Video Transformers Reorganize Dynamic Collision Cues Across Depth

Research Question 1 asks how dynamic collision cues and representation geometry change across depth before compression. Figure 2 combines layerwise probes with pooled-feature and token-level measures, allowing task discrimination, physical dynamics, and within-clip structure to be compared on the same backbone.

#### 4.1.1 Dynamic variables are most linearly accessible at intermediate depth

Linear probes on pooled block representations, cross-validated with grouping by video, show two opposing depth profiles (Figure 2A–C; Table 2). Low-level temporal change is already highly accessible in the earliest representation: temporal-difference energy reaches $R^2 = 0.978$ at Layer 0. More structured physical dynamics are most accessible at intermediate depth. Flow magnitude peaks at Layer 4 ($R^2 = 0.960$), as do approach versus retreat (AUC = 0.881), residual radial looming ($R^2 = 0.623$), and ego-expansion rate ($R^2 = 0.708$). Collision-label discrimination follows the opposite profile, increasing from AUC = 0.691 at Layer 0 to a maximum of 0.815 at Layer 11. Thus, the layer that makes relational motion easiest to decode is not the layer that best separates collision from non-collision videos.

Table 2. Layerwise accessibility of dynamic and collision-related variables

| Target | Peak layer | Peak value | Value at L11 |
|---|---|---|---|
| Temporal-difference energy ($R^2$) | L0 | 0.9780 | 0.8243 |
| Flow magnitude ($R^2$) | L4 | 0.9596 | 0.8524 |
| Approach versus retreat (AUC) | L4 | 0.8814 | 0.8373 |
| Residual radial looming ($R^2$) | L4 | 0.6228 | 0.4889 |
| Ego-expansion rate ($R^2$) | L4 | 0.7077 | 0.5748 |
| Collision label (AUC) | L11 | 0.8147 | 0.8147 |

#### 4.1.2 Representation dimensionality and token structure are axis-specific

The dimensionality analyses also revealed that across-video feature diversity and within-video token diversity evolve differently (Figure 2D–E). Pooled-feature effective rank, which measures variation among clip-level representations across different videos, increased substantially from early to later layers (from 3.1 to a maximum of 85.9 at Layer 10). However, within-clip token effective rank peaked around the early-to-intermediate blocks (97.9 at Layer 3) and declined toward the end of the network (49.2 at Layer 11). A model can therefore become richer in its across-video summary while its tokens within any one clip become less diverse. Conflating the two would make late layers look either uniformly redundant or uniformly rich, and neither is correct.

The axis-specific similarity analysis (Figure 2F) revealed an even clearer structure. Adjacent temporal-token cosine similarity increased from approximately 0.92 in the early layers to 0.970 at Layer 8, indicating that representations from neighbouring temporal bins remain highly similar as processing progressed. Spatial similarity however changed more dramatically and fell to approximately 0.297 at Layer 11. This suggests that VideoMAEv2 representations can become highly redundant across time while retaining spatial differentiation, even as their pooled clip-level representations remain discriminative across videos. This measured asymmetry motivates the controlled token-compression experiment below: if redundant information is concentrated more strongly along the temporal axis, merging temporally adjacent tokens should be safer than an otherwise matched spatial merge.

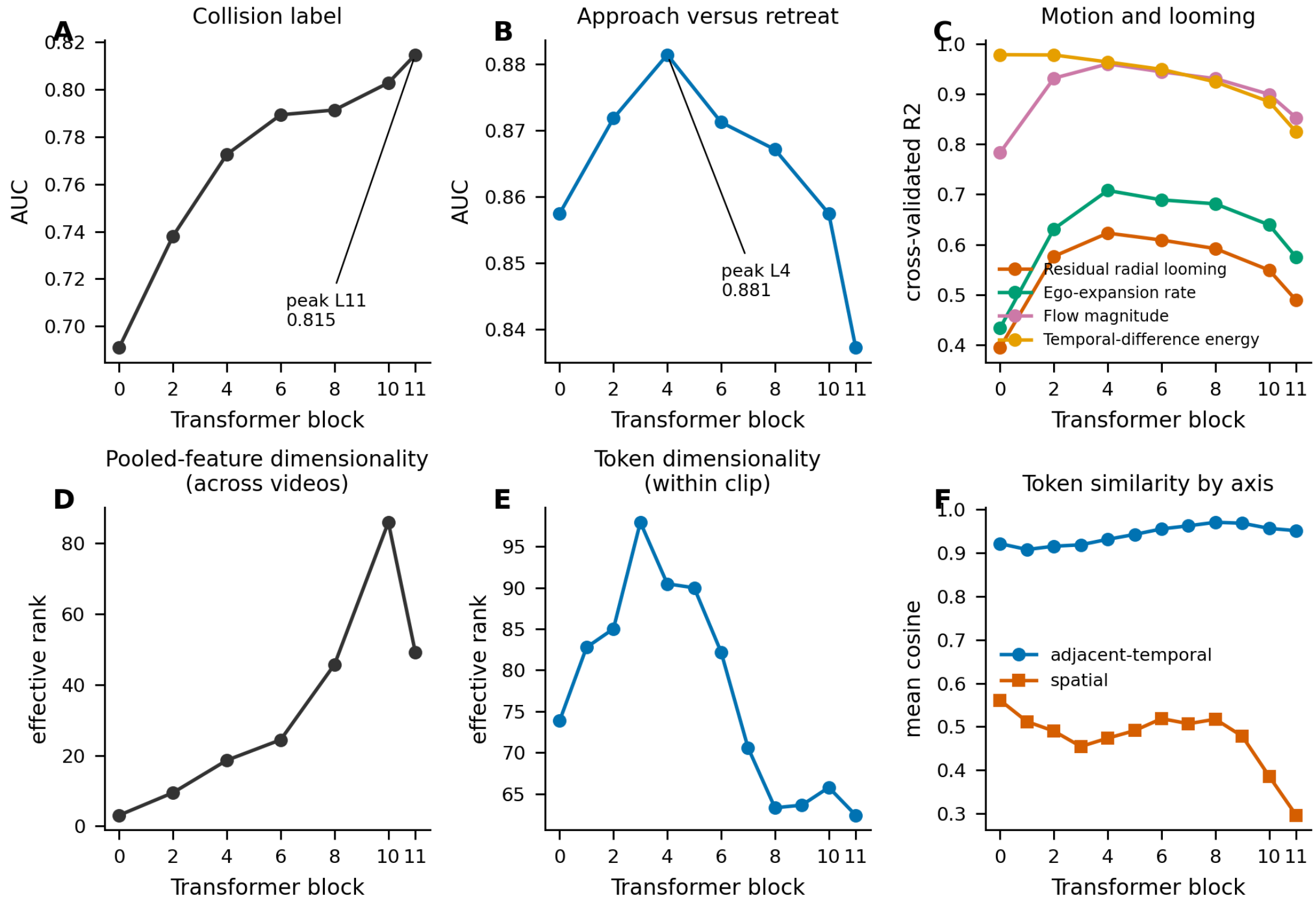

Figure 2. Evolution of dynamic information and token redundancy across VideoMAEv2 depth. (A) Collision-label discrimination strengthens in later blocks. (B) Approach versus retreat peaks at intermediate depth. (C) Flow, looming, expansion, and temporal-difference probes. (D) Pooled-feature effective rank across clips. (E) Token effective rank within clips. (F) Adjacent temporal and spatial cosine similarity. Probes measure linear accessibility, not causal use by the classifier.

### 4.2 Axis-Specific Token Compression

#### 4.2.1 Indiscriminate token removal is harmful at matched compute

We next intervene on the token stream at approximately twofold backbone-FLOP compression (356.99 to about 178.50 GFLOPs). Figure 3A and C and Tables 3, 5, and 6 summarize the measured latency and paired mean-AP changes. The random controls show that the transformer does not simply tolerate losing half of its token computation regardless of which tokens are retained. With the MLP-768 readout, fixed-schedule random removal reduces mean AP by -0.0269 [-0.0514, -0.0034], and depth-aware random removal reduces mean AP by -0.0417 [-0.0696, -0.0147]. Both confidence intervals exclude zero, showing that indiscriminate token removal causes a measurable loss in collision-prediction performance at approximately twofold compression. Structured methods show smaller point-estimate losses (Table 3), and none of their intervals excludes zero, indicating that no significant degradation was detected.

Table 3. Mean AP differences from dense inference at matched compute (MLP-768 head)

| Method | Selection | Schedule | Mean AP versus dense |
|---|---|---|---|
| R_fixed | random | fixed | -0.0269 [-0.0514, -0.0034] |
| R_depth | random | depth-aware | -0.0417 [-0.0696, -0.0147] |
| Tsp_depth | spatial merge | depth-aware | -0.0177 [-0.0496, +0.0128] |
| T_depth | temporal merge | depth-aware | -0.0102 [-0.0432, +0.0202] |
| V_depth | attention + merge | depth-aware | -0.0049 [-0.0324, +0.0212] |
| T_fixed | temporal merge | fixed | -0.0035 [-0.0323, +0.0242] |
| S_fixed | redundancy drop | fixed | -0.0025 [-0.0348, +0.0267] |

#### 4.2.2 Temporal and spatial compression behave differently

The matched-axis contrast in Figure 3B tests the prediction from the similarity measurements. The schedule-only random comparator is analyzed separately so that a temporal advantage is not confused with the layer at which removal happens. $T_{depth}$ and $Tsp_{depth}$ share the same layer schedule, per-layer token counts, and estimated FLOPs; only the merge axis changes (see Tables 4 and 5).

$T_{depth}$ and $Tsp_{depth}$ share the same schedule, realized FLOPs, and per-layer token counts, differing only in whether merging runs along the temporal or spatial axis. On hard dynamic negatives (the stratum that most directly probes discrimination of genuinely dynamic scenes), temporal merging consistently outperforms spatial merging (Table 4). At $\rho = 0.60$, the temporal-minus-spatial hard-negative AUC difference is +0.0517 for the linear head and +0.0494 for MLP-768; at $\rho = 0.70$, the corresponding differences are +0.0607 and +0.0637. All eight point estimates across the four compression ratios and two heads favor temporal merging, although only four intervals exclude zero. The repeated directional pattern supports the conclusion that temporal and spatial redundancy are not interchangeable.

Table 4. Temporal versus spatial merging at matched schedules, FLOPs, and token counts (hard-negative AUC)

| **$\rho$** | **Linear head** | **MLP-768 head** |
|---|---|---|
| 0.50 | +0.0288 [-0.0188, +0.0737] | +0.0340 [-0.0174, +0.0821] |
| 0.60 | +0.0517 [+0.0070, +0.0943] | +0.0494 [+0.0009, +0.0953] |
| 0.70 | +0.0607 [+0.0153, +0.1069] | +0.0637 [+0.0134, +0.1150] |
| 0.80 | +0.0023 [-0.0199, +0.0234] | +0.0153 [-0.0117, +0.0421] |

Table 5. Token-compression compute and latency at $\rho = 0.50$ (MLP-768 head)

| **Method** | **Selection** | **Schedule** | **FLOPs ratio** | **GFLOPs** | **ms/clip** | **Tokens out** |
|---|---|---|---|---|---|---|
| dense | - | - | 1.00 | 356.99 | 12.69 | 1,568 |
| R_fixed | random | fixed | 0.50 | 178.50 | 6.68 | 224 |
| R_depth | random | depth-aware | 0.50 | 178.64 | 7.18 | 69 |
| Tsp_depth | spatial merge | depth-aware | 0.50 | 178.64 | 7.51 | 105 |
| T_depth | temporal merge | depth-aware | 0.50 | 178.64 | 7.51 | 105 |
| S_fixed | redundancy drop | fixed | 0.50 | 178.50 | 7.80 | 224 |
| T_fixed | temporal merge | fixed | 0.50 | 178.50 | 7.21 | 222 |

Table 6. Token-compression prediction outcomes at matched compute (MLP-768 head)

| **Method** | **Mean AP** | **AP at 1.5 s** | **Hard-neg AUC** | **Mean AP versus dense** |
|---|---|---|---|---|
| dense | 0.7478 | 0.6519 | 0.6412 | - |
| R_fixed | 0.7208 | 0.6444 | 0.6378 | -0.0269 [-0.0514, -0.0034] |
| R_depth | 0.7061 | 0.6448 | 0.5933 | -0.0417 [-0.0696, -0.0147] |
| Tsp_depth | 0.7301 | 0.6461 | 0.6017 | -0.0177 [-0.0496, +0.0128] |
| T_depth | 0.7376 | 0.6536 | 0.6357 | -0.0102 [-0.0432, +0.0202] |
| S_fixed | 0.7453 | 0.6415 | 0.6493 | -0.0025 [-0.0348, +0.0267] |
| T_fixed | 0.7443 | 0.6724 | 0.6780 | -0.0035 [-0.0323, +0.0242] |

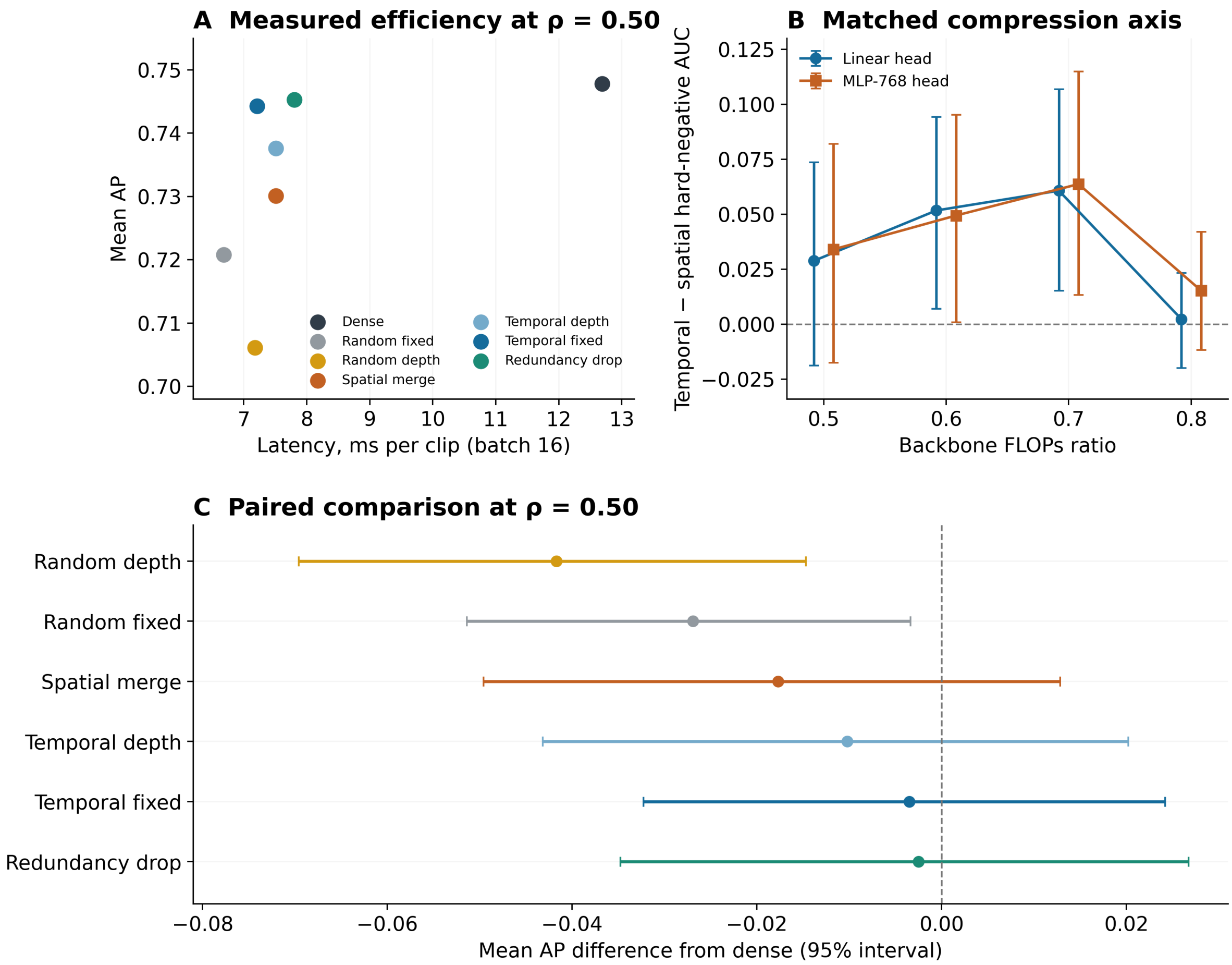

Figure 3. Token-compression performance at matched budgets. (A) Measured latency versus mean AP near half the dense backbone FLOPs. (B) Temporal-minus-spatial merging difference in hard-negative AUC under matched schedules, FLOPs, and token counts; error bars are paired 95% intervals. (C) Mean-AP difference from dense inference at ρ = 0.50 with paired 95% intervals. Token similarity is reported separately in Figure 2F.

#### 4.2.3 Compression robustness is not explained by preservation of linearly decodable motion cues

The preceding analyses show that intermediate transformer layers contain strongly decodable motion information and that temporal and spatial token redundancy evolve differently across depth. We therefore ask whether preserving linearly decoded motion information explains which compressed method retains hard-negative performance. If structured compression worked by protecting the dynamic variables detected by the probes, probe preservation should separate the methods and predict robustness. The results show that neither condition holds.

Averaged over six probe targets and layers 4-11, the change relative to dense was -0.0048 for temporal merging, -0.0091 for depth-aware temporal merging, -0.0124 for redundancy dropping, -0.0128 for spatial merging, -0.0132 for random removal and -0.0179 for the attention-based merging. Temporal merging preserves probes best, but three of the five structured methods are level with or worse than random selection, and on the motion probes specifically random selection beat four of the five. Guided selection as a class does not preserve these probes better than chance selection.

Nor does preservation predict robustness. Across the six methods at $\rho = 0.50$, correlations with hard-negative AUC were r = +0.218 for mean probe preservation and r = +0.041 for token effective rank. The method with the best-preserved effective rank (redundancy dropping, 59.2 of 63.2) was mid-table on hard negatives, and the method with the worst (attention-based, 11.7) was second best.

The only representation statistic that showed a strong association with hard-negative performance was residual temporal redundancy in the late layers. Residual adjacent-temporal cosine at Layer 11 against hard-negative AUC gave Spearman $\rho_s$ = -0.886 across the six methods, and remained close to -0.90 after excluding the most degenerate method. This is consistent with the matched temporal-versus-spatial intervention reported above, in which temporal merging was safer than spatial merging under equivalent computational budgets.

A further result rules out another simple interpretation. Temporal merging does not protect spatial differentiation. It reduces it, raising spatial cosine at Layer 11 from 0.2877 to 0.4809. All merging methods do this and all dropping methods leave spatial cosine near dense. This effect is expected mechanically because averaging tokens smooths them toward one another whichever axis it runs along. But it gives us a clear indication that the benefit of temporal merging arrives despite a loss of spatial diversity, not because spatial diversity was protected.

### 4.2.4 Computational savings are substantial except for memory

We next examine what the selected temporal-merging operating point provides in practical computational terms. It is worth noting that analytical FLOPs, measured clip latency, and peak memory answer different questions; a twofold reduction in analytical computation does not necessarily imply a twofold reduction in latency or memory use. Results for fixed temporal merging at $\rho = 0.50$ on the development split are summarized in Table 7.

Table 7. Efficiency and predictive performance of fixed temporal merging at $\rho = 0.50$

| | **Dense** | **T_fixed at $\rho = 0.50$** |
|---|---|---|
| Backbone GFLOPs | 356.99 | 178.50 (0.50x) |
| Latency, ms per clip at batch 16 | 12.69 | 7.21 (1.76x faster) |
| Peak GPU memory, MiB | 5,432 | 5,302 |
| Output tokens | 1,568 | 222 |
| Mean AP | 0.7478 | 0.7443 |
| AP at 1.5 s | 0.6519 +/- 0.0017 | 0.6724 +/- 0.0025 |
| Hard-negative AUC | 0.6412 | 0.6780 |

At $\rho = 0.50$, fixed-schedule temporal merging reduced estimated backbone computation from 356.99 to 178.5 GFLOPs, corresponding to a 50% reduction in transformer computation. Measured latency at batch size 16 decreased from 12.69 to 7.21 ms per clip, equivalent to approximately a 1.76x speedup. The number of output tokens decreased from 1,568 to 222.

Mean AP changed only slightly, from 0.7478 to 0.7443, while AP at the 1.5-s anticipation horizon increased from 0.6519 to 0.6724 and hard-negative AUC increased from 0.6412 to 0.6780, although their corresponding uncertainty intervals include zero.

The effect on memory is however much smaller. Peak GPU memory decreases only from 5,432 to 5,302 MiB. This may arise from that compression begins after block 2. The first three blocks always run at the full 1,568 tokens, so at least approximately 25% of the dense transformer FLOPs must still be incurred. Latency for the attention-scored arm is excluded from this comparison because its implementation recomputes attention for scoring rather than reusing the block own attention.

### 4.3 Importance-Guided MLP Pruning

The MLP experiment tests a different source of redundancy from token merging. Figure 4A and B compare guided and random unit retention, while Figure 4C–E examines whether poorer fixed-threshold performance reflects damaged ranking or changed calibration.

#### 4.3.1 Importance-guided MLP pruning reveals structured capacity redundancy

Scoring MLP units by a first-order Taylor criterion and retaining the top fraction per block (Figure 4A–B; Table 8) produces a retention curve with a distinctive shape. Guided pruning preserves discriminative ranking far below the retention level at which random removal collapses.

Table 8. AUC under importance-guided and random MLP pruning

| MLP units retained | Parameter fraction | Guided AUC | Random AUC (3 seeds) |
|---|---|---|---|
| 1.0 | 1.000 | 0.7925 | - |
| 0.8 | 0.8686 | 0.7845 | 0.7507 +/- 0.0249 |
| 0.5 | 0.6715 | 0.7533 | 0.5285 +/- 0.0416 |
| 0.4 | 0.6058 | 0.7218 | 0.4777 +/- 0.0347 |
| 0.3 | 0.5401 | 0.6412 | 0.4782 +/- 0.0368 |
| 0.2 | 0.4743 | 0.4158 | 0.4245 +/- 0.0745 |

The dense model achieved an AUC of 0.7925. Retaining 80% of MLP units reduced guided AUC only slightly to 0.7845, and retaining 50% still preserved an AUC of 0.7533. In contrast, random pruning at the same 50% retention level reduced AUC to 0.5285 $\pm$ 0.0416, close to chance. The difference became even larger at 40% retention. Importance-guided pruning retained an AUC of 0.7218, whereas random pruning fell to 0.4777 $\pm$0.0347. Only under more aggressive pruning did the guided model begin to collapse: by 0.2 both guided and random selection are at or below chance and indistinguishable. This reveals an importance-sensitive redundancy plateau followed by a capacity floor.

#### 4.3.2 Pruning changes calibration before it destroys ranking

A second observation changes how the degradation should be read (Tables 9 and 10). Pruning compresses the scale of the scores well before it destroys their ranking. Logit standard deviation falls from 2.509 at full retention to 0.839 at half retention and 0.121 at 0.2. Consequently, balanced accuracy evaluated at a fixed threshold of 0.5 appears to collapse (0.7121 to 0.5754 at half retention) while threshold-free ranking is nearly intact (AUC 0.7925 to 0.7533). Choosing the threshold on validation predictions only and applying it unchanged recovers most of the apparent loss (0.6769 at half retention), and expected calibration error falls from 0.236 to 0.064 after validation-only Platt scaling. Applied to the earlier full-versus-half comparison, this reduces the reported gap from 0.167 to 0.021 on the seen test set and from 0.143 to 0.029 on the unseen cohort.

Thus, much of the apparent degradation at moderate pruning levels is not caused by immediate destruction of the model's ranking ability. Instead, pruning first compresses the magnitude of the output scores and shifts their calibration. Only under stronger pruning does the underlying discriminative ordering itself deteriorate substantially. This implies that for collision-warning deployment, a decision threshold calibrated for the dense model cannot be assumed to remain valid after pruning, even when AUC or AP is largely retained. In other words, pruning introduces not only a capacity trade-off but also a threshold-calibration problem that should be reassessed before deployment.

Table 9. Structured MLP pruning: AP and AUC across retention levels. AP and AUC are averaged over the three anticipation horizons. Random AUC is reported as mean ± SD across three pruning seeds.

| MLP units retained | Parameter fraction | AP | AUC | Random AUC |
|---|---|---|---|---|
| 1.0 | 1.0000 | 0.7696 | 0.7925 | - |
| 0.8 | 0.8686 | 0.7706 | 0.7845 | 0.7507 +/- 0.0249 |
| 0.5 | 0.6715 | 0.7495 | 0.7533 | 0.5285 +/- 0.0416 |
| 0.3 | 0.5401 | 0.6581 | 0.6412 | 0.4782 +/- 0.0368 |
| 0.2 | 0.4743 | 0.4462 | 0.4158 | 0.4245 +/- 0.0745 |

Table 10. Threshold and calibration outcomes under importance-guided MLP pruning

| MLP units retained | BA at 0.5 | BA at validation threshold | ECE raw | ECE after Platt | Logit SD |
|---|---|---|---|---|---|
| 1.0 | 0.7121 | 0.7158 | 0.1575 | 0.0688 | 2.509 |
| 0.8 | 0.6951 | 0.7089 | 0.1059 | 0.0710 | 1.853 |
| 0.5 | 0.5754 | 0.6769 | 0.2361 | 0.0636 | 0.839 |
| 0.3 | 0.5510 | 0.5925 | 0.1626 | 0.0638 | 0.517 |
| 0.2 | 0.4894 | 0.4992 | 0.0709 | 0.0339 | 0.121 |

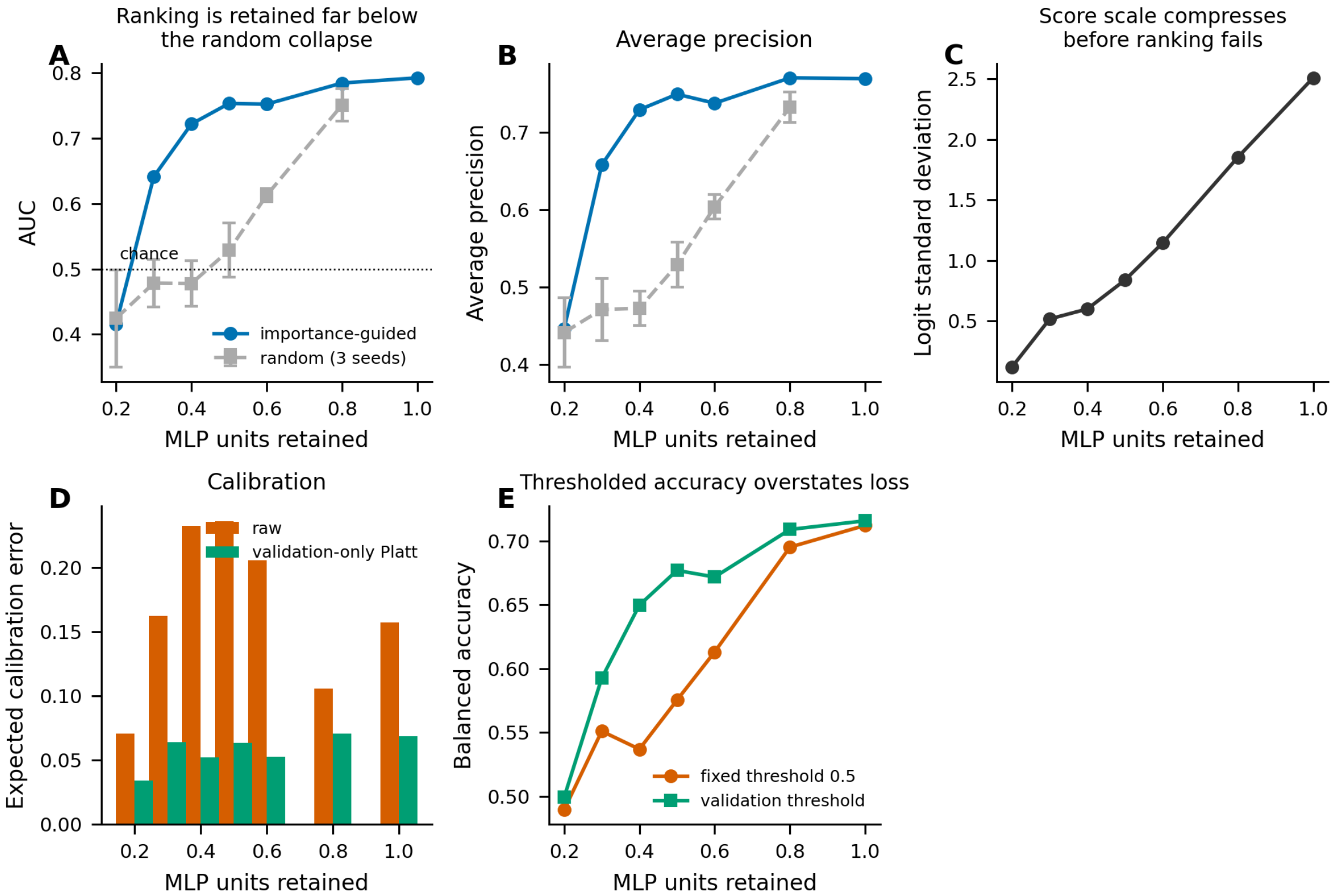

Figure 4. Importance-guided MLP-unit suppression and score calibration. (A, B) AUC and AP for guided versus random unit retention. (C) Logit-scale compression. (D) Expected calibration error before and after validation-only sigmoid scaling. (E) Balanced accuracy at fixed and validation-selected thresholds. Error bars for random retention summarize three seeds. Unit-retention ratios differ from estimated whole-model parameter fractions.

## 5 Conclusions

This study explores the spatiotemporal redundancy in the VideoMAEv2 model for collision anticipation in videos. It serves as a reference for worker–equipment proximity and struck-by hazard anticipation, echoing the broader adoption of transformer-based models in the research community for collision prediction, but this study examines a very underexplored question in terms of the pursuit of better-performing models with higher computational requirements, and of the redundancy in network capacity and token representation. This work aims to set up the stage where we can still retain the established research progress in different algorithms and architecture designs, while the compression methods trim the model to a smaller size without sacrificing its performance. Although this study only tests the VideoMAEv2 model, the pruning and token-merging methods remain universal and can easily be extended to many different architectures, and we expect this could become a new paradigm for model performance reporting in future studies.

Empirically, we found in the VideoMAEv2 model that depth-resolved probes show how information flows across layers: low-level temporal change is accessible early, physically interpretable motion variables are most linearly accessible at intermediate depth but diminish in the later layers, where collision-label discrimination becomes stronger. Controlled token interventions show that redundancy is axis-specific rather than simply depth-dependent. Late adjacent-temporal summaries were much more similar than distinct spatial tokens. Matched temporal and spatial merging have different effects on hard dynamic negatives; at $\rho = 0.50$, fixed temporal merging halves the estimated backbone FLOPs and reduces measured latency from 12.69 to 7.21 ms per clip. Importance-guided MLP pruning substantially outperforms random unit removal, while pruning also changes logit scale and calibration before discriminative ranking collapses. Recalibration partly recovered thresholded performance, but the MLP hook experiment has not yet demonstrated a corresponding hardware speedup. These findings add collision-anticipation-specific evidence to video-token-efficiency studies. The findings are consistent with previous studies that work on video understanding tasks, such as DyCoke (Tao et al. 2025), and with studies that explicitly emphasize separating temporal and spatial information, such as TimeSformer (Bertasius et al. 2021). Future research could leverage this and explore different token-merging methods developed in the video understanding domain, such as EVAD, which combines a dense keyframe with sparse temporal context (Chen et al. 2023); FrameFusion, which merges across time at corresponding spatial positions (Fu et al. 2025); and VAGNet, which preserves detailed spatial encoding first and then performs cheaper long-range temporal reasoning (Vipulananthan and Chitraranjan 2026) to collision anticipation in the video stream.

Another emerging direction future research is encouraged to explore is how the transformer architecture echoes or differs from visual processing in the human brain. Some recent precedent studies have explored representational similarity and layerwise brain alignment, asking whether the hierarchy inside the transformer corresponds to the temporal or anatomical hierarchy in the human brain (Lahner et al. 2024; Small et al. 2024; Goldstein et al. 2025; Cai et al. 2026), which can give a more granular understanding and a potential explanation of why particular blocks work or not, or why redundancy exists.

**Acknowledgement**

The author acknowledges the use of the Apollo GPU cluster, a service provided by the Sydney Informatics Hub, a Core Research Facility of the University of Sydney.